\pdfoutput=1

\documentclass[11pt]{article}

\usepackage{EMNLP2022}

\usepackage{times}
\usepackage{latexsym}
\usepackage{graphicx}
\usepackage{subcaption}
\usepackage{multirow}

\usepackage[T1]{fontenc}

\usepackage[utf8]{inputenc}

\usepackage{microtype}

\usepackage{inconsolata}

\title{Privacy Adhering Machine \textit{Un-learning} in NLP }

\author{Vinayshekhar Bannihatti Kumar,
   Rashmi Gangadharaiah,  
  Dan Roth \\
  AWS AI Labs\\
   \texttt{\{vinayshk,rgangad,drot\}}@amazon.com\\
  }

\begin{document}
\maketitle
\begin{abstract}
Regulations introduced by General Data Protection Regulation (GDPR) in the EU or California Consumer Privacy Act (CCPA) in the US have included provisions on the \textit{right to be forgotten} that mandates industry applications to remove data related to an individual from their systems. In several real world industry applications that use Machine Learning to build models on user data, such mandates require significant effort both in terms of data cleansing as well as model retraining while ensuring the models do not deteriorate in prediction quality due to removal of data. As a result, continuous removal of data and model retraining steps do not scale if these applications receive such requests at a very high frequency. Recently, a few researchers proposed the idea of \textit{Machine Unlearning} to tackle this challenge. Despite the significant importance of this task, the area of Machine Unlearning is under-explored in Natural Language Processing (NLP) tasks. In this paper, we explore the Unlearning framework on various GLUE tasks \cite{Wang:18}, such as, QQP, SST and MNLI. We propose computationally efficient approaches (SISA-FC and SISA-A) to perform \textit{guaranteed} Unlearning that provides significant reduction in terms of both memory (90-95\%), time (100x) and space consumption (99\%) in comparison to the baselines while keeping model performance constant.

\end{abstract}

\section{Introduction}
The use of user generated content in building ML-based solutions in many industry applications is quite common. As new data is acquired, these ML models undergo constant updates to further improve the performance. Such data can contain sensitive information such as names, account ids, email addresses and so on. Typically, user generated content goes through several redaction systems to remove such sensitive information before being used for retraining purposes. Industry applications also adopt rigorous approval processes with opt-in or opt-out capabilities to ensure that user generated content is used in a safe manner while raising awareness among users on how their generated content will be used.

More recently, the GDPR and CCPA provisioned the right to be forgotten, introducing new regulations that mandates industry applications to support deletion of user generated content when requested by users. Although privacy concerns have gained significant importance now, the scale at which these models are built make it extremely expensive and time-consuming to remove data efficiently while providing \textit{complete removal guarantees}.

Unlearning assures the users that their data has been completely removed when they prefer to have their data erased \cite{Lucas:2019}. This task is different from Differential Privacy that focuses on models that ensure that the users' information in the training data cannot be inferred and provides guarantees on the effect of an individual examples. Such a task is challenging as removal of data points can have significant deterioration on the performance of the models \textit{aka catastrophic unlearning} \cite{Min:19, Golatkar:19, Nguyen:20, Nguyen:22}. 

Approaches that guarantee complete removal of users' data points have recently gained momentum \cite{Lucas:2019, Tam:22} but under explored on NLP tasks. Motivated by SISA (\textbf{S}harded, \textbf{I}solated, \textbf{S}liced and \textbf{A}ggregated) training \cite{Lucas:2019} applied to CV datasets, we explore an approach that trains multiple ML models in isolation on disjoint shards and its slices. Model checkpoints are saved after training each slice in a shard. When a request for deletion is received, the corresponding datapoint is deleted from its slice and the model checkpoint upto the datapoint is used to further retrain the model. Although the SISA framework can be applied to any learning algorithm, it is still impractical in NLP settings as storing model checkpoints that include large language models (such as, BERT \cite{Devlin:18}) is both space and time consuming. We propose extensions to this SISA framework (1) (\textbf{SISA-FC}) requires storing only task-specific layers (2) (\textbf{SISA-A}) uses Adapters \cite{Houlsby:19} that requires storing only the adapter weights. These extensions prevent storing the entire model checkpoints, hence reducing time, memory and space footprints. We further improve upon the approaches by creating shards such that the least number of slices are affected when requests for removal are made, further reducing the time required to retrain the models. 
To the best of our knowledge this paper makes the first attempt to explore the task of Unlearning on various NLP tasks. The contributions of this paper are summarized as follows:
\begin{itemize}
    \item we explore the task of Unlearning on various NLP tasks (GLUE tasks such as QQP, SST and MNLI)
    \item we explore SISA-FC and SISA-A for NLP-based models that do not require storing large model checkpoints, thereby significantly reducing space, memory and time
    \item the paper proposes novel ways to partition the data, thereby reducing the number of slices affected and the retraining time while ensuring least degradation in the overall model quality.
\end{itemize}
The paper is organized as follows. Section \ref{sec:rel} briefly describes some of the related work in the area of Unlearning. Section \ref{sec:data} describes the NLP datasets used to evaluate the approaches proposed in this paper. Section \ref{sec:model} explains the proposed approaches (SISA-FC and SISA-A) followed by results in Section \ref{sec:results}. We finally conclude and provide future extensions to this work in Section \ref{sec:conclude}.

\section{Related Work}
\label{sec:rel}
Privacy preserving ML provides guarantees on bounds ensuring the contribution from the data points is as low as possible \cite{Sarwate:2018, Abadi:2016, Jang:22, Yu:21, Li:21}. Machine \textit{Unlearning} on the other hand focuses on \textit{complete} removal of training examples, ensuring that there is \textit{zero} contribution of the training samples to the model. As a result, Unlearning assures the users that their data has been completely removed when they prefer to have their data erased \cite{Lucas:2019}. A simple approach to forget training samples would be to re-train the models after removal of the examples. Such an approach is not scalable and computationally expensive when systems receive removal requests at high frequencies. 

Machine Unlearning has gained significant attention in industry applications as a means to allow users to completely delete their data from ML models \cite{Aman:21,Nguyen:22,Thomas:20}. Regulations introduced by regulatory bodies apply to all forms of user generated content. 
However, most of the recent approaches tackle this problem of Unlearning in Computer Vision (CV) settings \cite{Ronak:22, Golatkar:19, Lucas:2019}. It is imperative to explore unlearning strategies on textual data that can contain user sensitive information or personally identifiable information (PII). 

Our work is inspired by SISA training \cite{Lucas:2019} which was originally applied to CV datasets. The framework provides a strategic way to limit the influence of a data point in the training procedure. The approach trains models in isolation on disjoint shards created by partitioning the training data. When a request for removal is made, only the affected model is retrained. Shards are further broken down into Slides to decrease the time required to unlearn. During inference, they use ensemble strategies to aggregate predictions of individual models. We extend this framework for NLP models. Our paper also proposes approaches to partition the data such that the least number of shards/slices are affected, thereby further reducing the re-training time.

\section{Datasets}
\label{sec:data}
Glue~\cite{Wang:18} tasks are divided into three categories namely , "SINGLE-SENTENCE TASKS", " SIMILARITY AND PARAPHRASE TASKS" , " INFERENCE TASKS". We pick "SST-2" dataset from the first, "QQP" dataset from the second and "MNLI" dataset from the third respectively. In order to test the generalizability of our approach in terms of performance, time taken to retrain models and the memory they occupy on disk, we pick the first $60,000$ samples across each datasets and use 20\% of this as our test split. Having the same number of training and evaluation samples helps us to compare training times across different types of NLP tasks. For each of the datasets, we use accuracy to measure model performance. The tasks are explained below: \\
\textbf{SST-2:} This is a movie review corpus from ~\citet{socher2013recursive} that consists of a sentence form the review along with the sentiment associated with that sentence.\\
\textbf{QQP:} Quora Question Pair corpus~\cite{Wang:18} consists of two questions and a label indicating if the two questions are duplicates of each other. \\
\textbf{MNLI:} The MNLI~\cite{williams2017broad} corpus is built of top of SNLI. This consists of a premise, hypothesis and a label indicating if the premise and hypothesis are in entailment, contradiction or they are neutral. \\

\begin{figure*}
		\includegraphics[width=2\columnwidth]{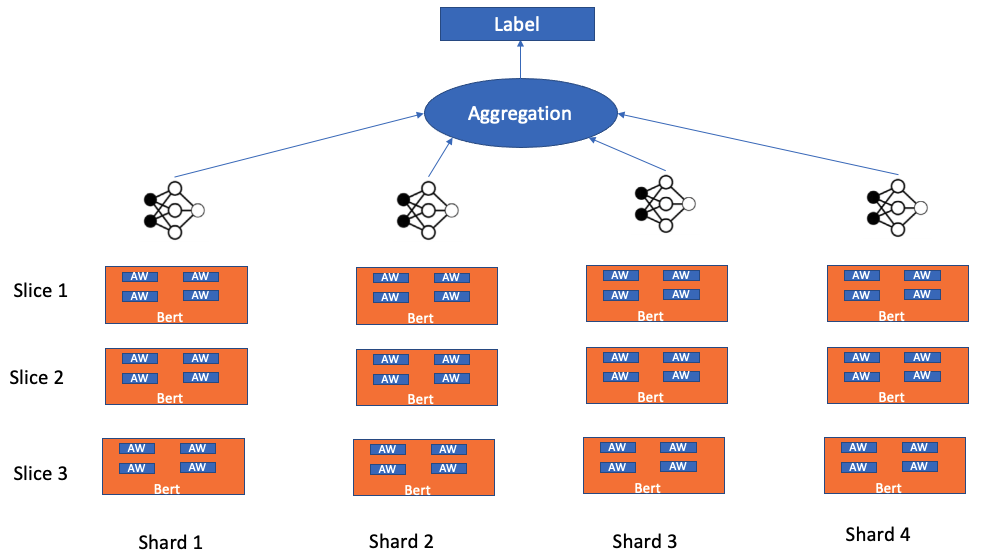}
		\caption{Architecture of SISA-A which uses S slices and R shards. One model is built for each shard and the labels are aggregated using a majority voting strategy.}
		\label{fig:model_arch}
	\end{figure*}

\section{Models}
\label{sec:model}
The simplest way to make a model forget the datapoint it has seen during training is to remove the datapoint from the training set and re-train the model. However, this is computationally very expensive for large models and we need an efficient way to re-training models if we want to forget a datapoint. ~\citet{Lucas:2019} presented Sharded, Isolated, Sliced and Aggregated (SISA) training approach in order to "un-learn" a datapoint. While they measured the performance of this approach on computer vision datasets, no work has looked at its performance on NLP datasets. We explain the algorithm and our approach to make it parameter efficient for NLP specific models.  

\subsection{SISA}
Figure~\ref{fig:model_arch} shows the working of the SISA algorithm. The entire training dataset is split into $S$ shards. Each shard is made up of $R$ slices. There is one model that is trained for every shard. In order to begin training, we can pick any ML model and then use gradient descent to train the model. While training, the model goes through the data, slice by slice saving a checkpoint after training for each slice. Finally once the model finishes training on the final slice, the model is saved and mapped to the shard. This process if continued for all the shards. During inference, each model belonging to a shard predicts the label and the labels are aggregated similar to model ensembling. When an un-learning request comes in, we pick the shard where this data point is present, then go to the slice that contains the un-learning request. We delete the datapoint from this slice, take the checkpoint that was trained up until that checkpoint and continue training the model. This guarantees that the model forgets the un-learning request. 

\subsection{SISA modified with fully connected Layer (SISA-FC)}
While the SISA framework is flexible to be used with any model, in the NLP domain it is not practical to store a checkpoint after each slice. Moreover the pre-training/fine-tuning time of the full model will become a major shortcoming and we need a way to reduce both the training time and the memory footprint of these large models. A simple way to alleviate this short-coming is to use a base model and pre-train it on a generic corpus of text and then add fully connected layers on top of it. Only the parameters from the linear layers are fine tuned in the optimization process. This will reduce the overall training time as the backpropagation of gradients only happens in the final layers and also we will only need to store the weights of these additional parameters. 

\subsection{SISA modified with Adapters (SISA-A)}
While adding linear layers to SISA might solves both the training time and the memory footprint issue, the raw performance of the model will take a hit when compared to fine tuning the entire model ~\cite{devlin2018bert}. One approach to keep the benefits of the linear layer while also optimizing for performance is to use Adapters~\cite{Houlsby:19} in the Encoder blocks of the transformer. While this increases the memory footprint of the model, it only accounts for about $1-5\%$ of the model parameters. Thereby providing us with $95-99\%$ memory benefits. 

\subsection{Baselines:}
We compare the performance of SISA-FC and SISA-A with respect to popular NLP settings. We fine-tune Bert on the same samples and show that there is not much impact on performance. We also show the majority classifier accuracy to show the improvement of SISA-FC and SISA-A against this baseline.

\subsection{Implementation details:}
We use the Bert-base model from Huggingface\footnote{https://huggingface.co/} to train on this task. We used a batch of 16, max length of 256, Adam~\cite{kingma2014adam} optimizer learning rate of $5e-3$ and train the model for 10 epochs when we are training the SISA-FC model. We also use the same hyper-parameters for SISA-A approach but add adapters using  Adapter-hub\footnote{https://docs.adapterhub.ml/training.html} and only update those parameters. We experiment with different slice sizes but keep the number of shards fixed at 5.

\section{Experiments and Results}
\label{sec:results}
For all the experiments in this paper we use a shard size of 5. 

\subsection{Evaluation Metrics:}
In order for an organization to allow customers to opt-out we need mechanisms that will enable re-training of models with low re-training time and memory foot-print while keeping the model performance as close to the original model as possible. Hence we look at the following three metrics : \\
\textbf{Accuracy:} Percentage of samples in the test set that the model predicts accurately. \\
\textbf{Re-training time:} The amount of time taken to delete the un-learning requests from the dataset and then re-train the model. \\
\textbf{Memory:} The amount of memory the final model takes up on disk.\\
We analyze the model across all the three metrics on $16$ uniformly randomly sampled requests unless stated otherwise. 

\subsection{Analysis of SISA-A}
\begin{figure*}[]
\begin{subfigure}{0.33\textwidth}
  \centering
  \includegraphics[width=\linewidth]{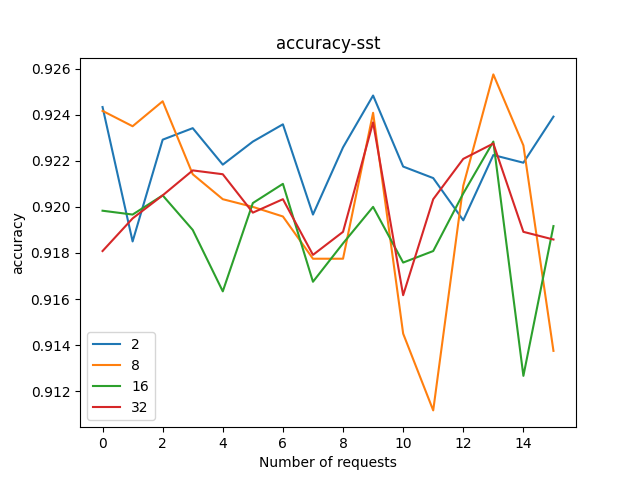}
  \captionof{figure}{SST}
  \label{fig:daily}
\end{subfigure}%
\begin{subfigure}{0.33\textwidth}
  \centering
  \includegraphics[width=\linewidth]{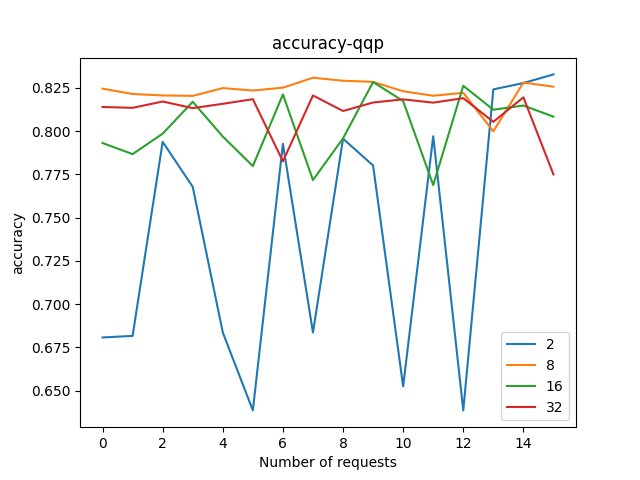}
  \captionof{figure}{QQP}
  \label{fig:mmf}
\end{subfigure}
\begin{subfigure}{0.33\textwidth}
  \centering
  \includegraphics[width=\linewidth]{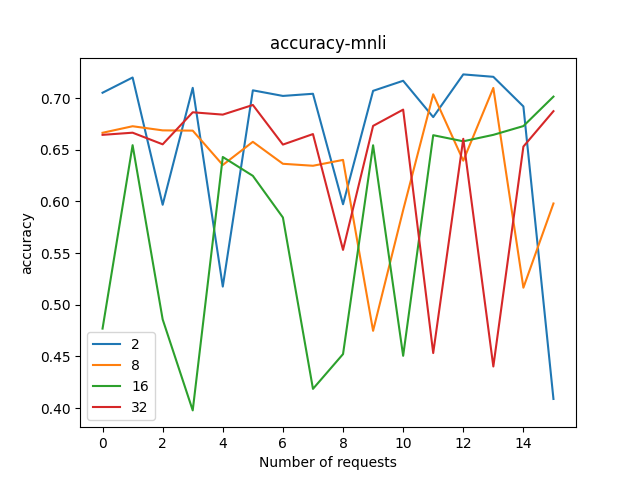}
  \captionof{figure}{MNLI}
  \label{fig:babi}
\end{subfigure}
\caption{Performance of SISA-A on the 3 datasets. \label{fig:exp1perf}}
\end{figure*}

Figure~\ref{fig:exp1perf} shows the performance of SISA-A for different slice sizes. We see that our approach is able to achieve the same level of performance as the BERT-base model even with the new setup we have with Adapters and SISA~\ref{table:baseline_comparision}. While it doesn't drop in accuracy by more than $1-6$\%, we see that the model has a lower training time of 100x. We also see that the memory occupied by the model is much lower than using the original SISA approach which would store the weights of the full model. We also see from Figure~\ref{fig:exp1perf} that the accuracy of our approach for slice size above 2 is almost the same. However with larger number of slices we see that the model has a much lower re-training time when compared to a slice size of two. This observation is consistent across all the 3 datasets. It is also to be noted that the re-training time does not increase linearly with more number of un-learning requests. The re-training time does increase with a positive delta. We show in Section~\ref{sec:pareto} on how we can make this re-training time flat across multiple un-learning requests. We also see that the model re-training time is constant across all the 3 datasets that were chosen.  

\begin{table*}[]
\small
\begin{tabular}{|c|cc|cc|p{2cm}|p{2cm}|}
\hline
\multirow{2}{*}{Dataset Name} & \multicolumn{2}{c|}{BERT-A}                          & \multicolumn{2}{c|}{SISA-A}                          & \multirow{2}{2cm}{Accuracy Reduction (\%)} & \multirow{2}{2cm}{Re-training time gain (\%)} \\ \cline{2-5}
                              & \multicolumn{1}{c|}{Accuracy} & Re-training time (s) & \multicolumn{1}{c|}{Accuracy} & Re-training time (s) &                                          &                                             \\ \hline
SST                           & \multicolumn{1}{c|}{0.93}     & 1580                 & \multicolumn{1}{c|}{0.92}     & 147                  & 1                                        & 9748                                        \\ \hline
QQP                           & \multicolumn{1}{c|}{0.82}     & 1750                 & \multicolumn{1}{c|}{0.80}     & 145                  & 2                                        & 1106                                        \\ \hline
MNLI                          & \multicolumn{1}{c|}{0.75}     & 1868                 & \multicolumn{1}{c|}{0.70}     & 145                  & 6                                        & 1188                                        \\ \hline
\end{tabular}
\caption{We see that our SISA-A approach is only $1-6$\% worse when compared to the deletion baseline model but has a ~100x improvement in training time over the deletion baseline. }
\label{table:baseline_comparision}
\end{table*}

\begin{figure*}[]
\begin{subfigure}{0.33\textwidth}
  \centering
  \includegraphics[width=\linewidth]{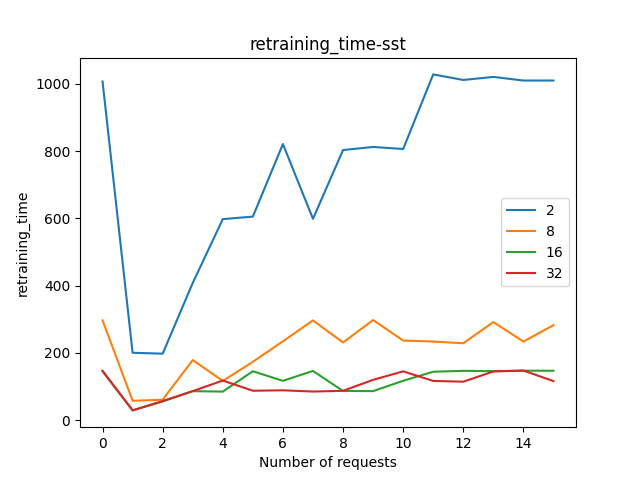}
  \captionof{figure}{SST}
  \label{fig:daily}
\end{subfigure}%
\begin{subfigure}{0.33\textwidth}
  \centering
  \includegraphics[width=\linewidth]{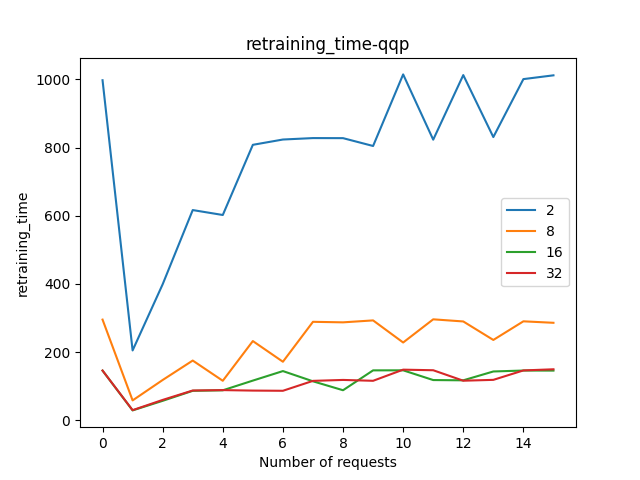}
  \captionof{figure}{QQP}
  \label{fig:mmf}
\end{subfigure}
\begin{subfigure}{0.33\textwidth}
  \centering
  \includegraphics[width=\linewidth]{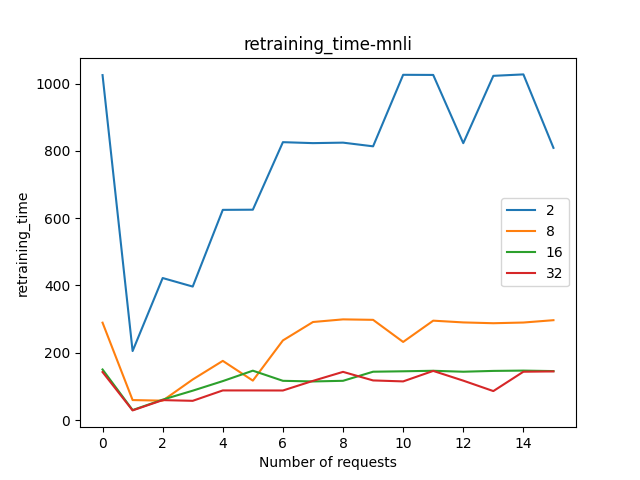}
  \captionof{figure}{MNLI}
  \label{fig:babi}
\end{subfigure}
\caption{Re-training time of SISA-A on the 3 datasets. \label{fig:exp1time}}
\end{figure*}

\subsection{Analysis of SISA-FC}
Figure~\ref{fig:exp4perf} shows the performance of SISA-FC for different slice sizes. We observe that the model performance for different un-learning requests is lower than that of SISA-A by $20-30$\% on all the tasks. This is due to the fact that there are far lesser number of weights that were used to fine-tune the model. However in low memory settings this approach would work better than the SISA-A approach as it occupies lesser memory. We also observe that the re-training time to just train the final layer is much lower than that of SISA-A. But other observations made with respect to SISA-A still holds with SISA-FC.

\subsection{Comparison Between SISA-FC and SISA-A}
\begin{figure*}[]
\begin{subfigure}{0.33\textwidth}
  \centering
  \includegraphics[width=\linewidth]{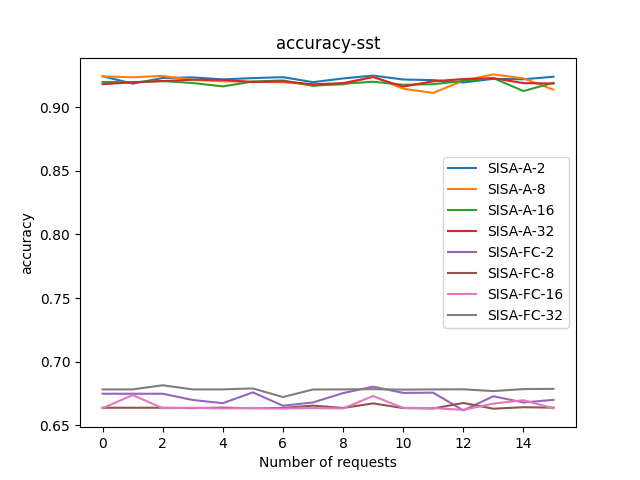}
  \captionof{figure}{SST}
  \label{fig:daily}
\end{subfigure}%
\begin{subfigure}{0.33\textwidth}
  \centering
  \includegraphics[width=\linewidth]{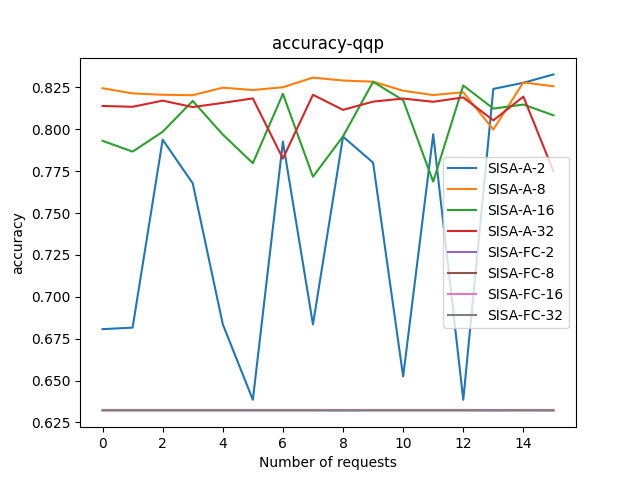}
  \captionof{figure}{QQP}
  \label{fig:mmf}
\end{subfigure}
\begin{subfigure}{0.33\textwidth}
  \centering
  \includegraphics[width=\linewidth]{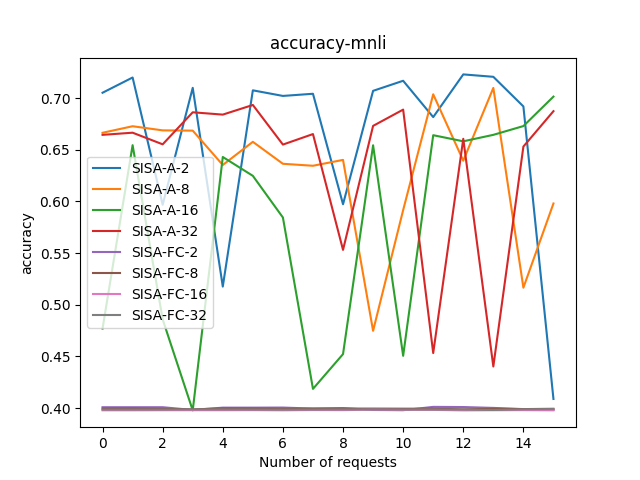}
  \captionof{figure}{MNLI}
  \label{fig:babi}
\end{subfigure}
\caption{Performance of SISA-FC on the 3 datasets. \label{fig:exp4perf}}
\end{figure*}

\begin{figure*}[]
\begin{subfigure}{0.33\textwidth}
  \centering
  \includegraphics[width=\linewidth]{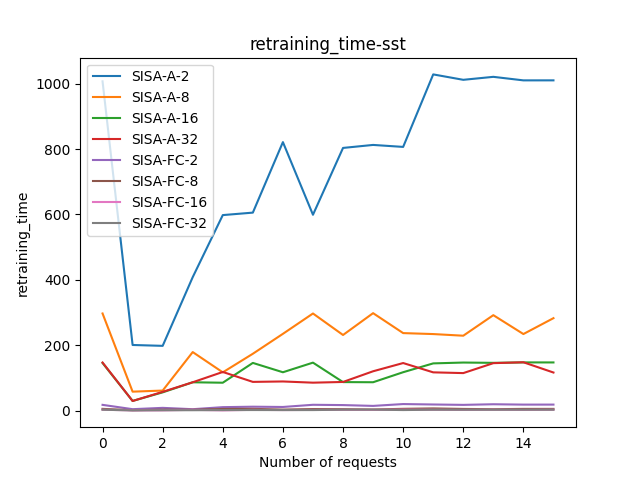}
  \captionof{figure}{SST}
  \label{fig:daily}
\end{subfigure}%
\begin{subfigure}{0.33\textwidth}
  \centering
  \includegraphics[width=\linewidth]{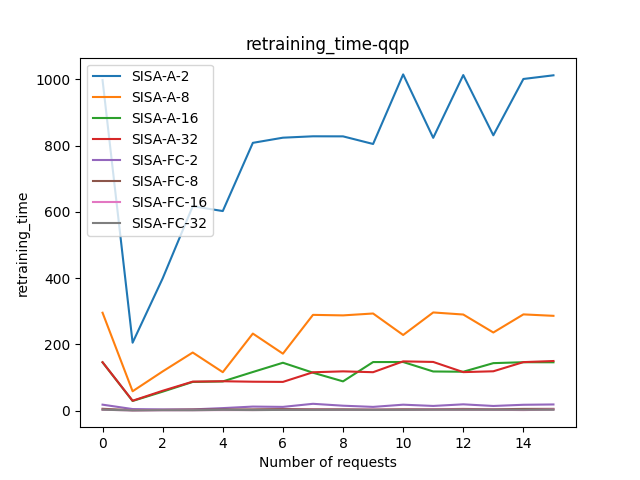}
  \captionof{figure}{QQP}
  \label{fig:mmf}
\end{subfigure}
\begin{subfigure}{0.33\textwidth}
  \centering
  \includegraphics[width=\linewidth]{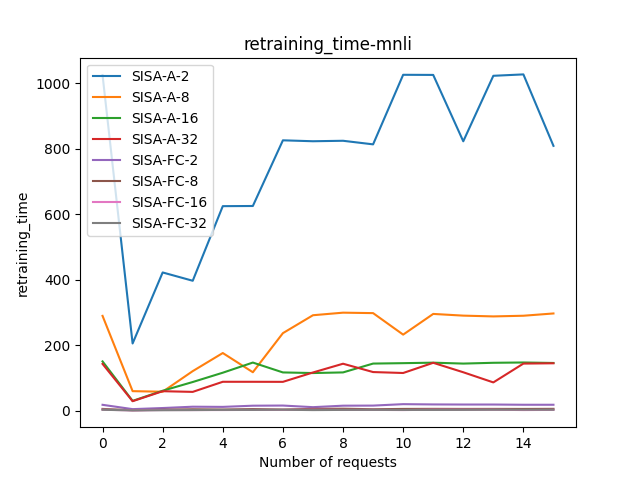}
  \captionof{figure}{Accuracy}
  \label{fig:babi}
\end{subfigure}
\caption{Re-training time of SISA-FC on the 3 datasets. \label{fig:exp4perf}}
\end{figure*}

From Figures~\ref{fig:exp4perf}, we see that the SISA-A approach significantly outperforms the SISA-FC approach in terms of accuracy. Across different slice sizes we see that we can get an absolute gain of 20-30\% in accuracy depending on the task. While the original adapter paper does note this performance difference, it is not that significant. When we apply adapters to SISA we see that the performance gain is much higher.  

\subsection{Profiling requests based on probability of occurrence}
\label{sec:pareto}
In the above two experiments, we assumed that the un-learning requests are uniformly randomly chosen from all of the training points. However, in a practical setting this might not be the case. Organizations can group customers into high risk and low risk customers based on their probability of opting out. In order to simulate such a scenario, we use the Pareto distribution. Pareto distribution is defined using the equation below:

\begin{equation}
    p(x) = a.m^{a} / x^{a+1}
\end{equation}

\begin{figure*}[]
\begin{subfigure}{0.33\textwidth}
  \centering
  \includegraphics[width=\linewidth]{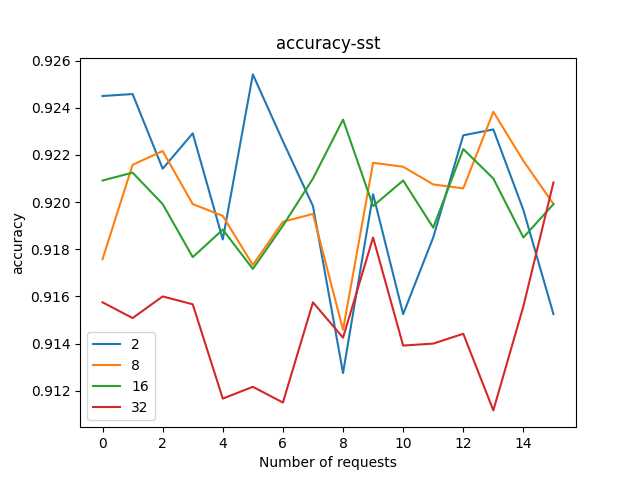}
  \captionof{figure}{SST}
  \label{fig:daily}
\end{subfigure}%
\begin{subfigure}{0.33\textwidth}
  \centering
  \includegraphics[width=\linewidth]{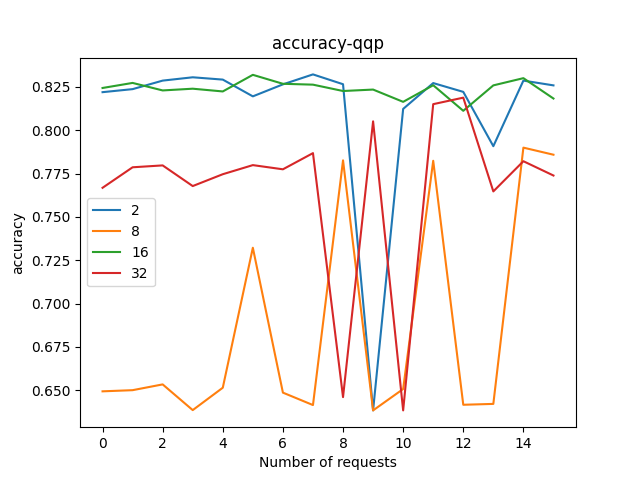}
  \captionof{figure}{QQP}
  \label{fig:mmf}
\end{subfigure}
\begin{subfigure}{0.33\textwidth}
  \centering
  \includegraphics[width=\linewidth]{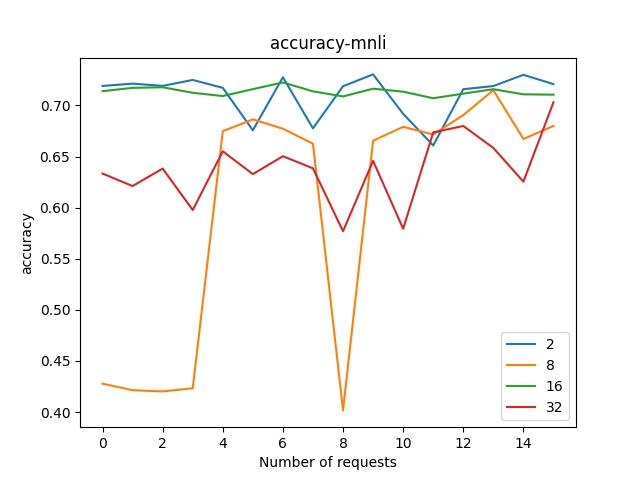}
  \captionof{figure}{MNLI}
  \label{fig:babi}
\end{subfigure}
\caption{Accuracy of SISA-A on Pareto distribution. \label{fig:exp3accuracy}}
\end{figure*}

\begin{figure*}[htb!]
\centering
\begin{subfigure}{0.30\textwidth}
  \centering
  \includegraphics[width=\linewidth]{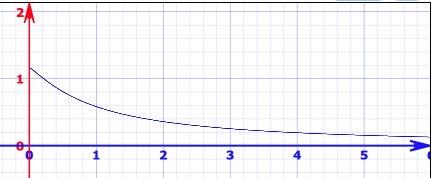}
  \captionof{figure}{Pareto Distribution}
  \label{fig:daily}
\end{subfigure}%
\begin{subfigure}{0.30\textwidth}
  \centering
  \includegraphics[width=\linewidth]{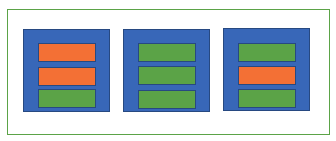}
  \captionof{figure}{Slices affected with Pareto in orange}
  \label{fig:mmf}
\end{subfigure}
\begin{subfigure}{0.30\textwidth}
  \centering
  \includegraphics[width=\linewidth]{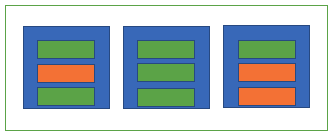}
  \captionof{figure}{Slices affected with Inverse Pareto in orange}
  \label{fig:babi}
\end{subfigure}
\caption{Pareto distribution and the shards it affects \label{fig:pareto}}
\end{figure*}

When we plot the Pareto distribution we get the following curve. 80\% of the mass is in the head of the distribution and 20\% in the tail. When we sample requests based on this distribution we get data-points belonging to the shards shown in Figure~\ref{fig:exp3accuracy}. For the purpose of this experiment we set $m$ to $1$ and $a$ to $1.16$ However if we take the mirror image of the Pareto distribution and sample from that distribution then the shards that will be affected is shown in Figure~\ref{fig:pareto}. We experiment with both these types of sampling to see its affect on model performance and re-training time. 

\begin{figure*}[]
\begin{subfigure}{0.33\textwidth}
  \centering
  \includegraphics[width=\linewidth]{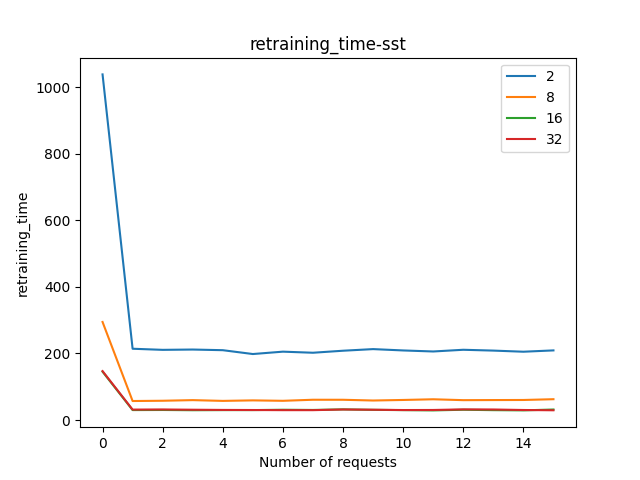}
  \captionof{figure}{SST}
  \label{fig:daily}
\end{subfigure}%
\begin{subfigure}{0.33\textwidth}
  \centering
  \includegraphics[width=\linewidth]{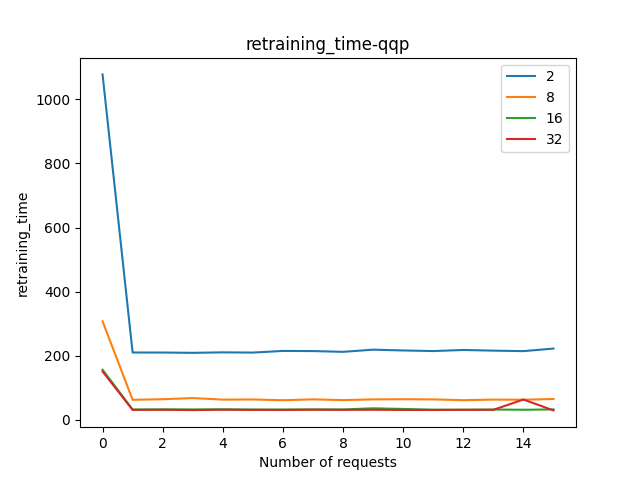}
  \captionof{figure}{QQP}
  \label{fig:mmf}
\end{subfigure}
\begin{subfigure}{0.33\textwidth}
  \centering
  \includegraphics[width=\linewidth]{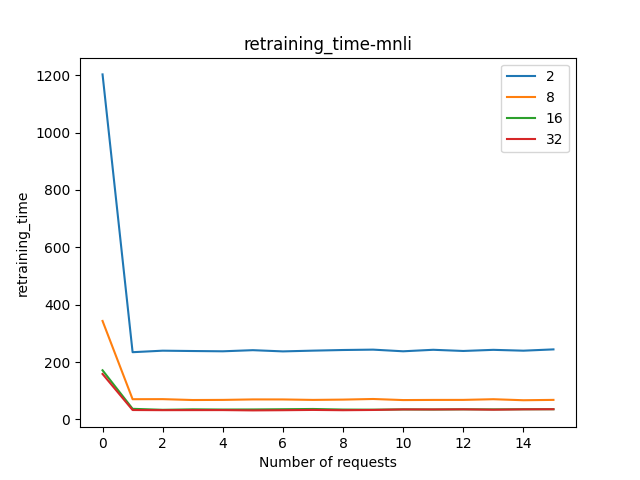}
  \captionof{figure}{Accuracy}
  \label{fig:babi}
\end{subfigure}
\caption{Re-training time of SISA-A on Inverse Pareto distribution. \label{fig:exp3timeinv}}
\end{figure*}

\begin{figure*}[]
\begin{subfigure}{0.33\textwidth}
  \centering
  \includegraphics[width=\linewidth]{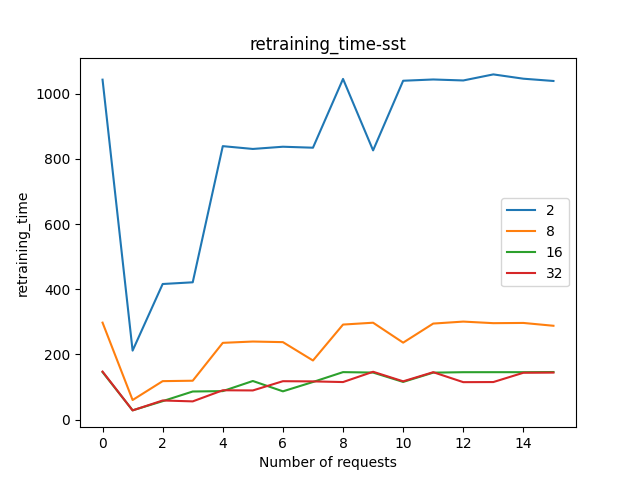}
  \captionof{figure}{SST}
  \label{fig:daily}
\end{subfigure}%
\begin{subfigure}{0.33\textwidth}
  \centering
  \includegraphics[width=\linewidth]{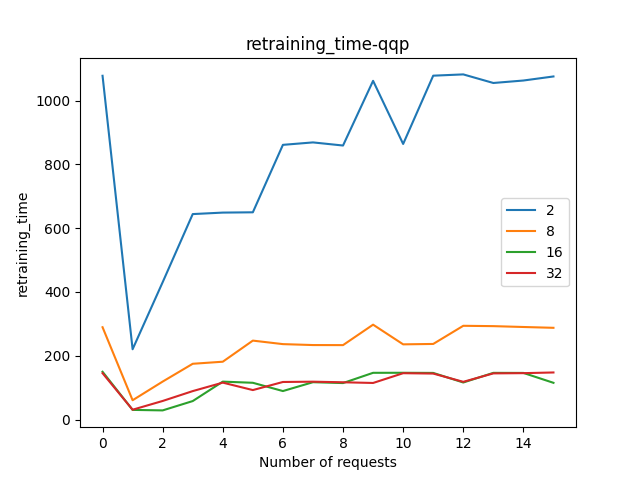}
  \captionof{figure}{QQP}
  \label{fig:mmf}
\end{subfigure}
\begin{subfigure}{0.33\textwidth}
  \centering
  \includegraphics[width=\linewidth]{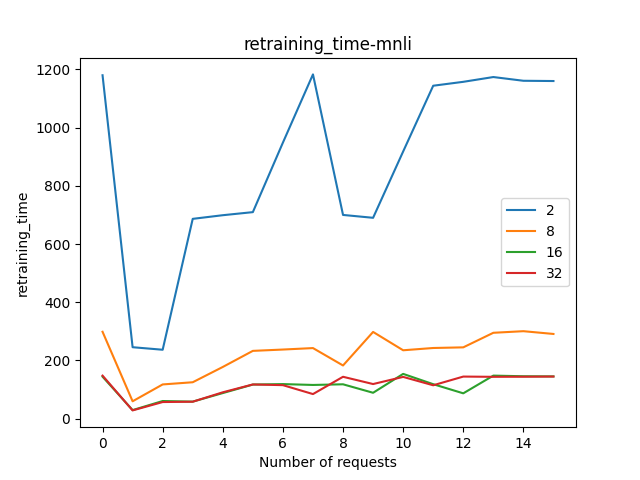}
  \captionof{figure}{MNLI}
  \label{fig:babi}
\end{subfigure}
\caption{Re-training time of SISA-A on Pareto distribution. \label{fig:exp3time}}
\end{figure*}

From Figures~\ref{fig:exp3time} and~\ref{fig:exp3timeinv}, we see that the training time of the inverse Pareto distribution is lesser than that of both the uniform distribution and Pareto distribution by 350\% and 360\% respectively. We also see that the re-training time of the inverse Pareto distribution is flat across multiple un-learning requests as they only affect the last few slices in the shard, thereby lowering training time. This opens up to possibilities of smarter customer segmentation while training. While we show one simple way to segment customers to gain on re-training time, we leave the possibilities of exploring other such strategies for future work.

\begin{figure*}[]
\begin{subfigure}{0.33\textwidth}
  \centering
  \includegraphics[width=\linewidth]{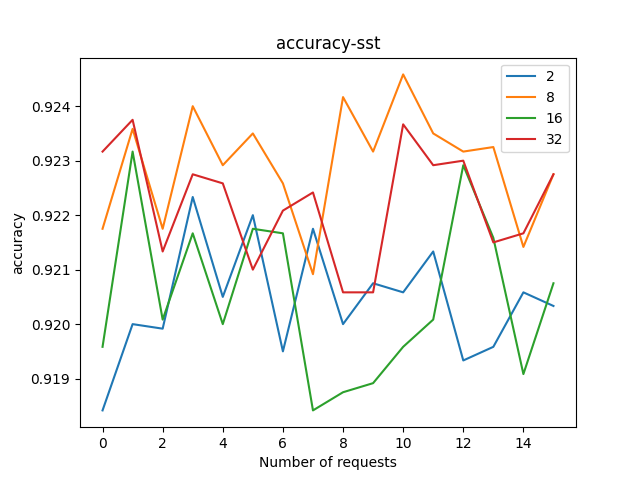}
  \captionof{figure}{SST}
  \label{fig:daily}
\end{subfigure}%
\begin{subfigure}{0.33\textwidth}
  \centering
  \includegraphics[width=\linewidth]{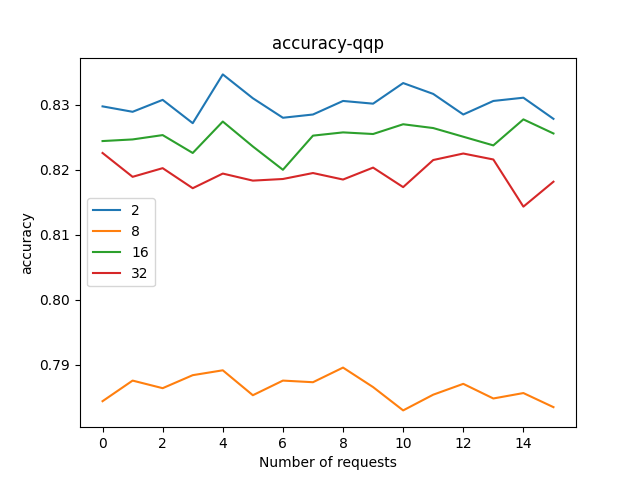}
  \captionof{figure}{QQP}
  \label{fig:mmf}
\end{subfigure}
\begin{subfigure}{0.33\textwidth}
  \centering
  \includegraphics[width=\linewidth]{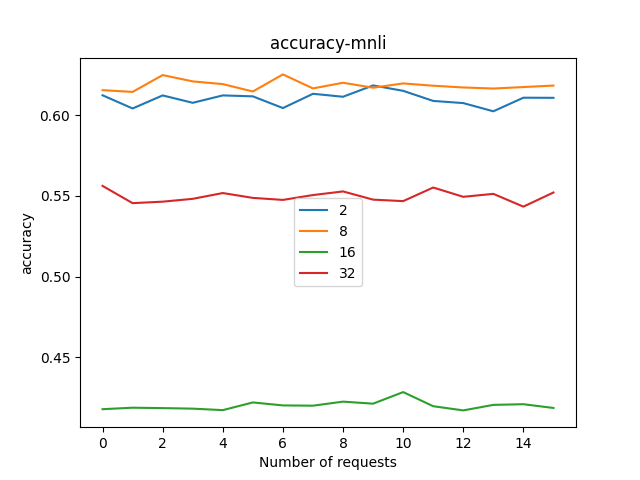}
  \captionof{figure}{MNLI}
  \label{fig:babi}
\end{subfigure}
\caption{Accuracy of SISA-A on Inverse Pareto distribution. \label{fig:exp3accuracyinv}}
\end{figure*}

\begin{figure*}[]
\begin{subfigure}{0.33\textwidth}
  \centering
  \includegraphics[width=\linewidth]{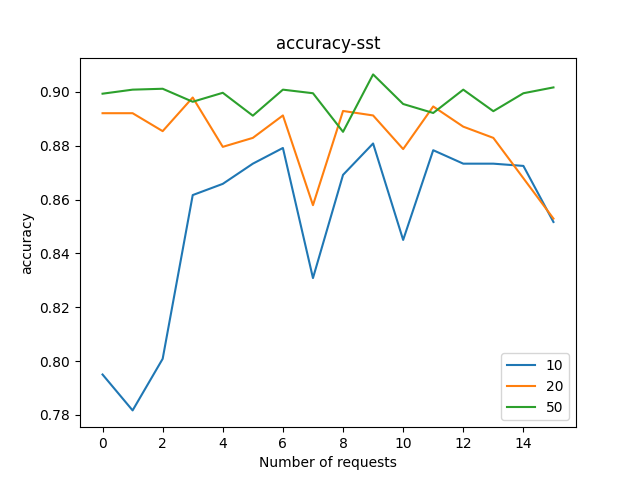}
  \captionof{figure}{SST}
  \label{fig:daily}
\end{subfigure}%
\begin{subfigure}{0.33\textwidth}
  \centering
  \includegraphics[width=\linewidth]{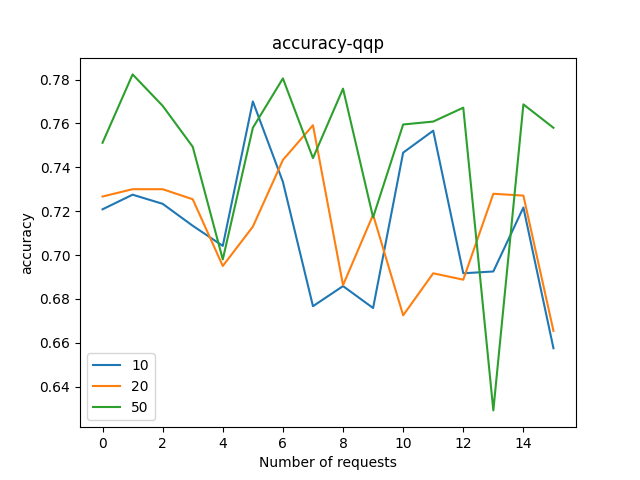}
  \captionof{figure}{QQP}
  \label{fig:mmf}
\end{subfigure}
\begin{subfigure}{0.33\textwidth}
  \centering
  \includegraphics[width=\linewidth]{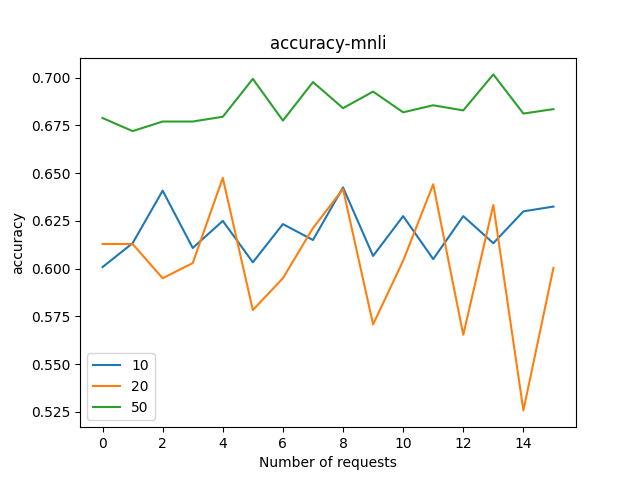}
  \captionof{figure}{Accuracy}
  \label{fig:babi}
\end{subfigure}
\caption{Accuracy of SISA-A on partial datasets. \label{fig:exp2accuracy}}
\end{figure*}

\begin{figure*}[]
\begin{subfigure}{0.33\textwidth}
  \centering
  \includegraphics[width=\linewidth]{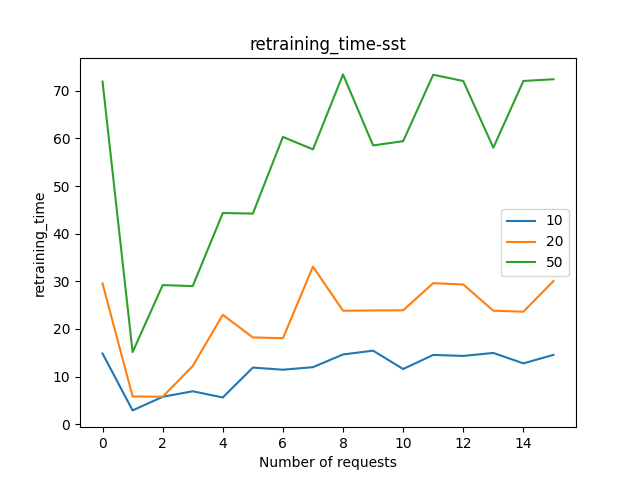}
  \captionof{figure}{SST}
  \label{fig:daily}
\end{subfigure}%
\begin{subfigure}{0.33\textwidth}
  \centering
  \includegraphics[width=\linewidth]{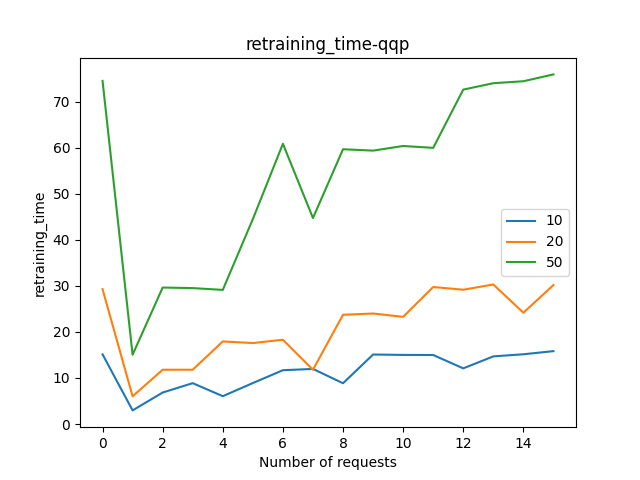}
  \captionof{figure}{QQP}
  \label{fig:mmf}
\end{subfigure}
\begin{subfigure}{0.33\textwidth}
  \centering
  \includegraphics[width=\linewidth]{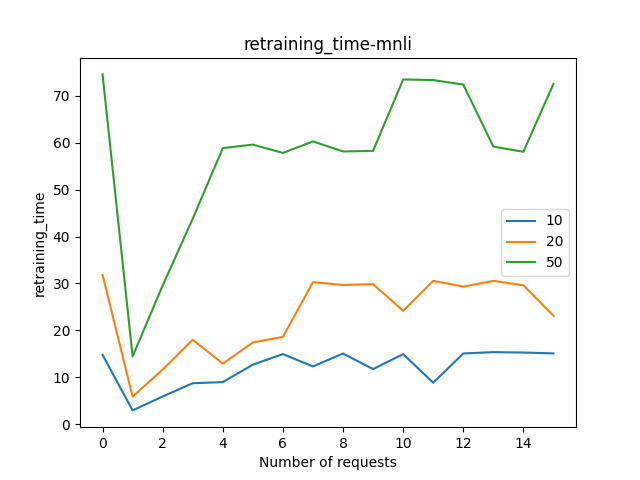}
  \captionof{figure}{Accuracy}
  \label{fig:babi}
\end{subfigure}
\caption{Re-training time of SISA-A on partial datasets. \label{fig:exp2time}}
\end{figure*}

\subsection{How does SISA-A perform on a portion of the data?}
In the next experiment, we look at the performance of SISA-A on portions of the training set to test the few shot generalizability of this approach. For the purpose of this experiment, we use 5 shards, 16 slices and uniform un-learning requests. We see that on all the different datasets, the performance in the few-shot setting does not match the performance of the model in the full-data setting. This is one of the major drawbacks of the SISA algorithm in low resource NLP settings. Since the model is forced to look at a portion of the data several times before it is check-pointed, the model tends to overfit on the slice it is training on. We hypothesize this to be the reason for lower model performance.

We also see that the BERT-base model performs much better than the SISA-A approach in the few-shot setting. While we show the short-coming of this approach in the few-shot setting we leave the further exploration to alleviate this issue for future work.

\subsection{Memory profiling}
\begin{figure*}[htb!]
\centering
\includegraphics[width=0.25\linewidth]{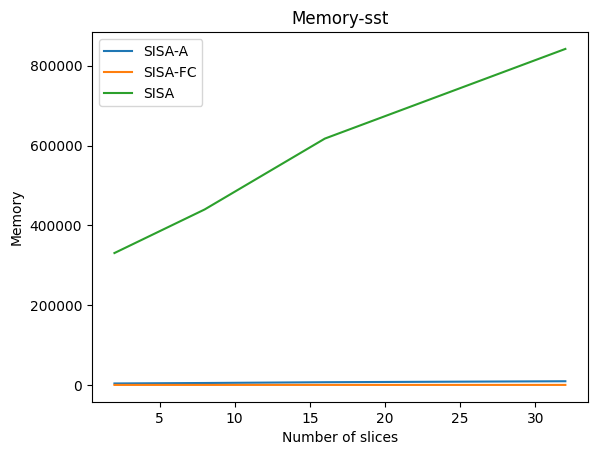}
\label{fig:daily}
\caption{Memory Occupied by the three approaches.\label{fig:memory}}
\end{figure*}

Figure~\ref{fig:memory} shows the memory occupied by the three different approaches compared in this work. We see that if we use the BERT model along with the vanilla SISA algorithm, the memory increases linearly as the number of slices increases. But with our approach, the memory increase is far less drastic (SISA-A and SISA-FC).

\section{Conclusion and Future Work}
\label{sec:conclude}
The central tenet of privacy regulations in the U.S and EU revolve around the concept of "notice and choice". This allows users to opt-out of data collection if they deem something violates their privacy needs. However, owing to the gap between technology experts and privacy regulators there has not been enough work around opting out of data that has already been used to train ML models. Through the SISA-A and SISA-FC approaches we show the working of the SISA algorithm in the NLP domain on Encoder based models. Because of the requirement of an aggregation layer in SISA we cannot extend this approach to the decoder based models and that is a limitation of our work. 

\pagebreak
\pagebreak

\bibliography{anthology,custom}
\bibliographystyle{acl_natbib}

\appendix

\section{Example Appendix}
\label{sec:appendix}

This is a section in the appendix.

\end{document}